\documentclass[letterpaper, 10 pt, conference]{ieeeconf}  

\IEEEoverridecommandlockouts                              

\usepackage{graphicx} 
\usepackage{subcaption}
\usepackage{hyperref}
\hypersetup{
    colorlinks=true,
    linkcolor=black,
    filecolor=black,      
    urlcolor=black,
    pdftitle={Overleaf Example},
    pdfpagemode=FullScreen,
    }
\usepackage{amsmath, mathtools, amssymb}
\usepackage{xcolor} 
\usepackage[inkscapelatex=false]{svg}

\usepackage{color, soul} 
\usepackage{booktabs}
\usepackage{multirow}
\usepackage[export]{adjustbox}
\usepackage{xspace}
\DeclareMathOperator*{\argmin}{arg\,min} 

\newcommand{\fig}[1]{Fig.~\ref{#1}}  
\newcommand{\tbl}[1]{Table~\ref{#1}}

\newcommand{\etal}[0]{{\em et al.~}} 
\newcommand{\eg}[0]{{\em e.g.,~}} 
\newcommand{\ie}[0]{{\em i.e.,~}}

\title{\LARGE \bf
Object Importance Estimation using Counterfactual Reasoning for Intelligent Driving
}

\author{Pranay Gupta$^{1}$, Abhijat Biswas$^{1}$, Henny Admoni$^{1}$, David Held$^{1}$
\thanks{$^{1}$Pranay Gupta, Abhijat Biswas, Henny Admoni and David Held are with the Robotics Institute, Carnegie Mellon University, Pittsburgh, PA, USA. {\tt\small pranaygu@andrew.cmu.edu, abhijat@cmu.edu, henny@cmu.edu, dheld@andrew.cmu.edu}. This work was supported by the CMU Argo AI Center for Autonomous Vehicle Research.}%
}

\begin{document}

\maketitle
\thispagestyle{empty}
\pagestyle{empty}

\begin{abstract}
The ability to identify important objects in a complex and dynamic driving environment is essential for autonomous driving agents to make safe and efficient driving decisions. It also helps assistive driving systems decide when to alert drivers. We tackle object importance estimation in a data-driven fashion and introduce HOIST - Human-annotated Object Importance in Simulated Traffic. HOIST contains driving scenarios with human-annotated importance labels for vehicles and pedestrians. We additionally propose a novel approach that relies on counterfactual reasoning to estimate an object's importance. We generate counterfactual scenarios by modifying the motion of objects and ascribe importance based on how the modifications affect the ego vehicle's driving. Our approach outperforms strong baselines for the task of object importance estimation on HOIST. We also perform ablation studies to justify our design choices and show the significance of the different components of our proposed approach.

\end{abstract}

\section{Introduction}

Assistive and autonomous driving agents need to reason about objects within the driving scene to make safe driving decisions. However, not all objects are equally important. The ability to identify important objects would allow autonomous driving systems to perform efficient path planning and resource management by reducing the number of objects that require tracking and modeling. Moreover, prior work has also shown that importance estimation can help reduce false positive alerts in assistive driving systems that detect driver inattention and preemptively warn the driver (\eg detecting a jaywalking pedestrian and highlighting them on a HUD)~\cite{wu2022toward}. Recent work has also shown improvement in autonomous driving agents' performance when contrastively trained to de-prioritize unimportant objects~\cite{biswas2022mitigating}. Thus, in this work, we focus on the task of important object identification in the context of driving.

In prior works~\cite{palazzi2018predicting, xia2019predicting, baee2021medirl, pal2020looking, fang2019dada} researchers have tried to identify salient regions in a driving scene by modeling driver attention through eye-gaze fixations. However, in the context of driving, the correlation between gaze and attention is questionable~\cite{kotseruba2021behavioral} (see Sec. 2) as the driver may not always look at important objects due to distractions~\cite{white2010blind}, inexperience~\cite{crundall2012some}, or paying attention via peripheral vision~\cite{biswas2023characterizing}. 
\begin{figure}[t]
  \centering
  
  \begin{subfigure}[t]{0.3\linewidth}
    \includegraphics[width=\linewidth]{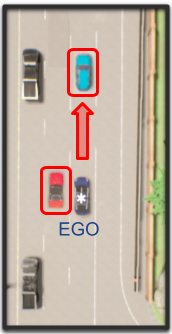}
    \caption{Red boxes indicate vehicles important for making safe driving decisions by the ego vehicle.}
    \label{fig:fig1a}
  \end{subfigure}
  \hfill
  \begin{subfigure}[t]{0.3\linewidth}
    \includegraphics[width=\linewidth]{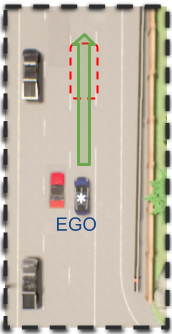}
    \caption{The blue car is important since its removal would cause the ego vehicle to speed up.}
    \label{fig:fig1b}
  \end{subfigure}
  \hfill
  \begin{subfigure}[t]{0.3\linewidth}
    \includegraphics[width=\linewidth]{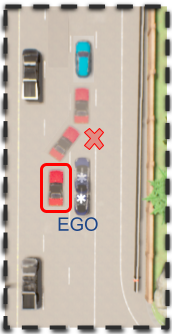}
    \caption{The red vehicle is important since its changing lanes would risk a collision with the ego vehicle.}
    \label{fig:fig1c}
  \end{subfigure}
  
  \setlength{\belowcaptionskip}{-15pt}
  \caption{We perform counterfactual reasoning to identify important vehicles in a driving scenario. We modify the motion of vehicles, and ascribe importance based on how the modification affects the ego vehicle's driving.}
  \label{fig:fig1}
\end{figure}

Another line of work~\cite{gao2019goal, zhang2020interaction, li2022important} tries to identify the importance of objects within the scene directly. However, the definition of ``important objects" used in that work is limited to objects which influence the driving behavior of the ego vehicle. We extend the definition of ``important objects" to include all the objects that a driver should be aware of, in order to drive the ego vehicle safely (Sec.~\ref{subsubsec:annotations}) \ie all objects necessary to maintain sufficient situational awareness~\cite{endsley1995toward}. This includes objects that both currently influence and \emph{can potentially influence} the driving decisions of the ego vehicle.  In addition to enhancing autonomous and assistive driving systems, identifying objects important according to our definition allows for the development of situational awareness support systems~\cite{wu2022toward}. 

Moreover, the object-importance datasets proposed in prior works~\cite{gao2019goal, zhang2020interaction} are unimodal (only ego-view RGB videos) and do not benefit from multiple sensor modalities (multi-view RGB cameras, LiDAR) for object detection and importance estimation; further, these datasets are not publicly accessible. To address these challenges, we introduce HOIST: Human-annotated Object Importance in Simulated Traffic. HOIST consists of driving scenarios from representative routes of the Longest6 benchmark~\cite{chitta2022transfuser} with human-annotated object importance. While prior works~\cite{gao2019goal, zhang2020interaction, li2022important} obtain annotations using ego-view driving videos, we collect annotations using bird's-eye-view RGB videos of the driving scene, as the bird's-eye-view offers more context regarding the driving scenario as compared to the front-view video. In addition to the RGB videos, HOIST also contains data from all sensor modalities and configurations supported by CARLA~\cite{dosovitskiy2017carla} (\eg  RGB Camera, LIDAR, GPS).

Our new dataset enables our second contribution: a novel approach for object importance estimation. We propose to determine an object's importance by answering counterfactual questions about those objects. Specifically, we modify the motion of objects in the scene and generate counterfactual scenarios. We gauge the  change in the ego vehicle's driving under the counterfactual (estimated using an autonomous driving model~\cite{chen2022learning}) to estimate each object's importance. 

We evaluate our novel approach on the task of object importance estimation, using the HOIST dataset . The results highlight the effectiveness of our approach in identifying important objects. We also perform extensive ablation studies, to validate our design choices and analyze the significance of the different components of our approach. Furthermore, since our approach relies on autonomous driving agents to predict the ego vehicle's behavior, analysing our failure cases can help identify driving scenarios where the autonomous driving agents behave unexpectedly. 

The main contributions of our work are:
\begin{itemize}
    \item HOIST, a novel dataset containing simulated driving scenarios with human-labeled important objects, which influence or can potentially influence the ego vehicle's driving. In addition to RGB videos, HOIST provides access to data from all the sensors configurations offered by CARLA~\cite{dosovitskiy2017carla}.
    \item A novel approach that employs counterfactual reasoning to estimate object importance in driving scenarios. We modify the motion of objects to generate counterfactual scenes and gauge the change in the ego vehicle's behavior to ascribe importance.
    \item Benchmarking and ablation experiments on HOIST that substantiate the effectiveness of our approach in detecting important objects in a driving scene. Our code and dataset will be available on our project page\footnote{\url{https://vehicle-importance.github.io}}.
\end{itemize}


\section{Related Works}
\label{sec:related}

\noindent\textbf{Object Importance Estimation in on-road driving:} The task of identifying important objects in the context of driving has garnered increasing attention from researchers. Prior approaches~\cite{gao2019goal, zhang2020interaction, li2020make} learn to predict object importance using end-to-end models, leveraging
the ego vehicle's goal~\cite{gao2019goal}, interactions between objects~\cite{zhang2020interaction}, or the ego vehicle's predicted behavior~\cite{li2022important} to identify important objects. 
Since manual annotation is time intensive, Li \etal also performs semi-supervised learning by generating ranking based pseudo-labels for the unlabeled data. Instead of learning end-to-end models for estimating object importance, our approach relies on counterfactual reasoning.

Another line of works~\cite{li2020make, li2023droid}, performs causal inference to identify risky objects. Similar to our approach, they remove objects and verify whether the removal produces a change in the ego vehicle's driving decision. While these works only consider removal, we estimate importance by perturbing other dynamic attributes of the objects. A third category of works attributes importance to objects using attention weights obtained from attention-based models trained for a different task, such as trajectory forecasting~\cite{vemula2018social, li2021rain, li2021spatio}, or end-to-end driving~\cite{Renz2022CORL, aksoy2020see}.

\noindent \textbf{Driver's Attention Prediction Methods:}
While driving, humans can selectively attend to relevant objects in the scene~\cite{xia2019driver, xia2020periphery}. Hence, prior works~\cite{palazzi2018predicting, xia2019predicting, baee2021medirl} tried to identify salient regions within a dynamic traffic scene by studying the driver's attention. Here, the driver's eye gaze serves as a proxy for the driver's attention. For example, Palazzi \etal\cite{palazzi2018predicting} proposes to use a multi-stream 3D convolutional network that leverages RGB frames,  optical flow, and segmentation maps to predict the driver attention heatmap. 
Baee \etal\cite{baee2021medirl} models eye-gaze fixations as a sequence of states and actions and proposes maximum entropy deep inverse reinforcement learning~\cite{ziebart2008maximum} to predict maximally rewarding fixation locations. In our work, we directly estimate the importance of objects within the scene, rather than detecting the driver's attention heatmaps.

\noindent \textbf{Driver's Attention Prediction Datasets:} Typically, methods for driver attention prediction leverage large-scale driver attention datasets comprising ego-centric driving videos with the corresponding driver gaze~\cite{xia2019predicting}. DReyeVE~\cite{palazzi2018predicting} was the first large-scale driver attention dataset recorded using eye-tracking glasses in a car. 
Recent driver attention datasets~\cite{fang2019dada, baee2021medirl} have focused on accident scenarios. Pal \etal\cite{pal2020looking} argues that driver gaze is insufficient in describing everything a driver should attend to, and proposes to use ground truth gaze map combined with segmentation masks of all the objects within the scene. Unlike these datasets, HOIST consists of data from a host of sensor observations (RGB cameras, LIDAR, GPS) along with explicitly annotated object importance, rather than using the driver gaze as a proxy for importance.


\section{HOIST: Human-annotated Object Importance in Simulated Traffic}

We consider the problem of object importance estimation in the context of driving. We define ``important objects" as the objects that the driver should watch out for in order to make safe driving decisions. Prior object importance estimation datasets~\cite{gao2019goal, li2022important} limit important objects to only those which currently influence the ego vehicle's driving decisions. Moreover, they only contain single-view RGB videos and  are not publicly accessible. Transcending these challenges, we curate \textbf{HOIST}: Human-annotated Object Importance in Simulated Traffic. HOIST is a new dataset of driving scenarios with human annotations of object importance. In addition to RGB videos, HOIST also provides access to data from a host of different types of sensors, including multi-view cameras and LIDAR. These different sensors will allow users to use this dataset to evaluate different methods of object importance estimation, as discussed in Sec.~\ref{sec:method}. 

\subsection{Dataset Curation}
\label{subsec:curation}
We obtain the driving videos for HOIST from the Longest6 benchmark~\cite{chitta2022transfuser}, which consists of routes from the CARLA~\cite{dosovitskiy2017carla} driving simulator. The Longest6 benchmark was introduced as a substitute for the official CARLA evaluation routes to overcome the compute restrictions set by CARLA. It consists of the six longest routes from each of the six towns in CARLA, resulting in a total of 36 routes. It preserves the most challenging aspects of the CARLA evaluation routes in terms of traffic density, weather conditions, and adversarial traffic scenarios. For our dataset, we select six representative routes (one from each town) of the Longest6 benchmark. We drive the ego vehicle through the six selected routes using a rule-based-expert~\cite{chitta2022transfuser}, recording simulation data along each route.  

We extract the bird's-eye-view videos from the simulation recording of each route and split them into smaller 13 second videos. Unfortunately, the routes in the Longest6 benchmark majorly involve driving in a straight lane. Driving in straight lanes leads to trivial cases for object importance estimation as the objects closely leading and following the ego vehicle are generally deemed important. To ensure that we include more diverse driving scenarios, we limit the number of videos with straight lanes and include videos where the ego vehicle is undergoing complex driving maneuvers like crossing intersections, turning, changing lanes, or stopping for pedestrians. We end up with a total of 409 videos.

\subsection{Human Annotations}
\label{subsubsec:annotations}
We hire annotators on Prolific\footnote{\url{https://www.prolific.co/}} to collect annotations for the videos obtained in the previous step (\ref{subsec:curation}). The annotators were first shown the bird's-eye-view RGB videos and were then asked to draw bounding boxes around important objects in the last frame of the video. In our instructions, we defined important objects as the ones the annotators would watch out for in order to drive the ego vehicle safely if they were to take control of the ego vehicle from the point at which the video ended. We instructed the annotators to focus on only two types of objects: vehicles and pedestrians \footnote{Details about the interface and the prompts used to obtain the annotations are available on our project page : \url{https://vehicle-importance.github.io}}. 

Our annotators reported that they are more than 18 years of age and have a valid US state-issued driving license. Each annotator did three practice rounds, two at the beginning and one at the end. These practice rounds were straightforward and were used to exclude annotators whose annotations demonstrated a lack of comprehension of the task. We hired 188 annotators and each annotated 15 videos. To mitigate the bias due to the driving habits of individual annotators, each video was annotated by at least 5 different annotators.

\subsection{Dataset Statistics}
\label{subsubsec:analysis}

The HOIST dataset consists of a total of 409 videos with  642 unique human-annotated important objects. Out of these 409 videos, 8.8\% of the videos have $0$ important objects, 44.9\%  have $1$ object, 32.2\%  have $2$ important objects, and 13.9\%  have more than $2$ important objects. In this work, we focus on identifying important vehicles and pedestrians. Out of the 642 total important objects, 597 are vehicles and 45 are pedestrians.


\section{Estimating Object Importance}
\label{sec:method}

In this section, we describe our novel method for estimating the importance of objects for making safe control decisions for the ego vehicle. Because the movement patterns of vehicles and pedestrians are different from each other, we propose to use different approaches to estimate their importance. For estimating the importance of vehicles, we perform counterfactual reasoning about the vehicles (Sec~\ref{subsec:vehicle_importance}). For pedestrians, we find their proximity to the ego vehicle to be a very strong cue for importance estimation (Sec~\ref{subsec:pedestrian_importance}). We explain both of these approaches in detail in the following sections. 

\begin{figure}[t]
  \centering
  \begin{subfigure}[t]{0.75\linewidth}
  \begin{subfigure}[t]{0.47\linewidth}
    \includegraphics[width=\linewidth]{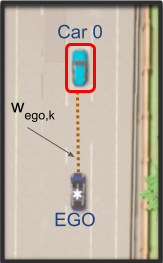}
    \caption{Original}
    \label{fig:fig2a}
  \end{subfigure}
  \hfill
  \begin{subfigure}[t]{0.47\linewidth}
    \includegraphics[width=\linewidth]{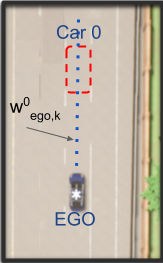}
  
    \caption{Counterfactual}
    \label{fig:fig2b}
  \end{subfigure}
  \end{subfigure}
  \setlength{\belowcaptionskip}{-15pt}
  \caption{The removal score is the L2 distance between the corresponding waypoints of the true trajectory (shown in a) and the predicted trajectory after removing Car 0 (shown in b).}
  \label{fig:fig2}
\end{figure}

\subsection{Vehicle Importance Estimation}
\label{subsec:vehicle_importance}
For estimating vehicle importance, our proposed approach performs counterfactual reasoning. We generate counterfactual scenarios by modifying the motion of the non-ego vehicles; we ascribe importance based on how the modifications affect the ego vehicle’s driving. We modify the ego vehicle's sensor observations to simulate modifications in the motion of the non-ego vehicles. We then compare the changes in the ego vehicle's trajectory under the counterfactual. If the ego vehicle's trajectory has significantly changed under the counterfactual, then the vehicle is deemed important (details below). We also perform collision estimation under the counterfactuals to ascribe importance. 

We leverage autonomous driving systems~\cite{chen2022learning} to predict the ego vehicle's trajectory. Typical autonomous driving systems~\cite{shao2023reasonnet, shao2022interfuser, wu2022trajectory, chen2022learning, chitta2022transfuser} perceive a scene through various sensor modalities (e.g. multiview RGB cameras and LIDAR) and predict the future trajectory for the ego vehicle. A trajectory is represented by a set of waypoints. We query such a system with the true and the counterfactual sensor observations and compare the change in the predicted waypoints. We base our importance predictions on two types of counterfactual scenarios: 

\begin{figure*}[t]
  \centering

  \begin{subfigure}[t]{0.32 \linewidth}
      \begin{subfigure}[t]{0.47\linewidth}
        \includegraphics[width=\linewidth]{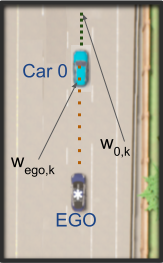}
        \caption*{Original}
      \end{subfigure}
      \begin{subfigure}[t]{0.47\linewidth}
        \includegraphics[width=\linewidth]{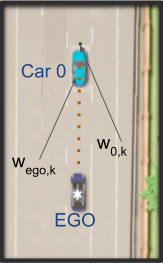}
        \caption*{Counterfactual}
      \end{subfigure}
  \caption{\textbf{Hard-Stop} perturbation.}      
  \label{fig:fig3a}
  \end{subfigure}
  \hfill
  \begin{subfigure}[t]{0.32 \linewidth}
      \begin{subfigure}[t]{0.47\linewidth}
        \includegraphics[width=\linewidth]{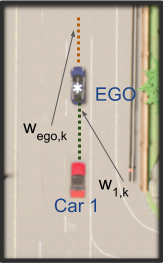}
        \caption*{Original}
      \end{subfigure}
      \begin{subfigure}[t]{0.47\linewidth}
        \includegraphics[width=\linewidth]{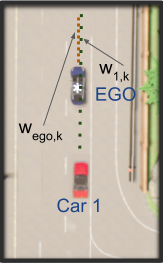}
        \caption*{Counterfactual}
      \end{subfigure}
      \caption{\textbf{Speed-Up} perturbation.}
      \label{fig:fig3b}
  \end{subfigure}
  \hfill
  \begin{subfigure}[t]{0.32 \linewidth}
      \begin{subfigure}[t]{0.47\linewidth}
        \includegraphics[width=\linewidth]{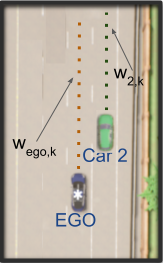}
        \caption*{Original}
      \end{subfigure}
      \begin{subfigure}[t]{0.47\linewidth}
        \includegraphics[width=\linewidth]{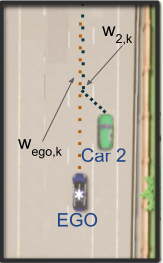}
        \caption*{Counterfactual}
      \end{subfigure}
      \caption{\textbf{Lane-Change} perturbation.}
      \label{fig:fig3c}
  \end{subfigure}
  \setlength{\belowcaptionskip}{-15pt}
  \caption{The velocity perturbation score assesses potential collision, when the non-ego vehicle undergoes a velocity perturbation. A collision occurs when the corresponding waypoints of the ego vehicle and the non-ego vehicle trajectories are within a threshold. The collisions occur at the $k^{th}$ waypoint. The trajectories of the ego vehicle and the non-ego vehicle are shown in in orange and green, respectively.}

  \label{fig:fig3}
\end{figure*}

\noindent \textbf{Removal:} To check if a vehicle affects the ego vehicle's driving decisions, we verify whether removing that vehicle from the scene would cause the ego vehicle to change its trajectory. We remove vehicles by modifying the sensor observations input to the autonomous driving system (See Section~\ref{subsubsec:implementation} for details). Consider the example shown in \fig{fig:fig2}. The importance of an object is calculated as the L2 distance between the waypoints predicted using the true sensor observations (\fig{fig:fig2a}) and the counterfactual observations (\fig{fig:fig2b}). We call this the Removal Score ($\mathcal{RS}$). Let $w_{ego, k}$ be the $k$th predicted waypoint of the ego vehicle with the true sensor observations, and let $\hat{w}_{ego, k}^{i}$ be the $k$th predicted waypoint of the ego vehicle under the counterfactual observation generated by removing the $i$th vehicle. The Removal Score $\mathcal{RS}_{i}$ for the $i$th vehicle is 
\begin{equation}
    \mathcal{RS}_i = \sum_{k \in \{1, 2, \ldots  K\}} || w_{ego, k} - \hat{w}_{ego, k}^{i} ||^{2}
\end{equation}
where $K$ is the total number of predicted waypoints.

\noindent \textbf{Velocity Perturbations:} The Removal Score identifies vehicles that affect the ego vehicle's trajectory in the current state. However, a vehicle can become a safety hazard if it suddenly changes its speed or direction. Thus, we check whether changing the velocity of a particular vehicle would make it susceptible to collisions with the ego vehicle (see \fig{fig:fig3}). We simulate velocity changes by perturbing the trajectories of the vehicles. Similar to the ego vehicle, we obtain the trajectories for non-ego vehicles using an autonomous driving agent. Recent methods for autonomous driving predict the trajectories of non-ego vehicles in the scene either by using deep models~\cite{chen2022learning} or by projecting it using a constant velocity model~\cite{shao2022interfuser}, where the constant velocity is calculated using a moving average over the past timesteps.
We consider three possible velocity perturbations. To identify vehicles that would become safety critical if they suddenly stopped, we generate a counterfactual scenario in which the vehicle makes a \textbf{hard stop}. To simulate hard stopping, we collapse all the waypoints of the vehicle into the first waypoint (See \fig{fig:fig3a}). Next, we identify the vehicles that would become safety critical if they suddenly \textbf{sped up}. For this, we generate a counterfactual scenario where we increase the vehicle's velocity. We simulate this by increasing the distance between consecutive waypoints of the vehicle's trajectory (See \fig{fig:fig3b}). Finally, to identify vehicles that would become safety critical if they suddenly \textbf{changed lanes}, we simulate lane change maneuvers for each object. We perturb the trajectories of the vehicles such that they first perform a 45-degree lane shift into an adjacent lane (left or right) and then follow a straight line (See \fig{fig:fig3c}). 

In addition to the vehicles in the scene, the ego vehicle might also have to deviate from its predicted trajectory under emergencies. Therefore, as a defensive driving measure, we also perturb the ego vehicle's trajectory according to these three velocity perturbations and similarly measure collisions with other vehicles under this counterfactual to estimate vehicle importance.

We consider a vehicle important due to velocity-based perturbations if it is predicted to collide with the ego vehicle. A collision is predicted if a waypoint from any of the non-ego vehicle's trajectories (predicted or perturbed) is within the safety threshold distance of a waypoint from the ego vehicle's trajectories (predicted or perturbed). Moreover, a predicted collision that happens sooner is considered more important than a predicted collision that happens later. Hence, we score vehicle importance based on how soon they are predicted to collide with the ego vehicle. Thus, the Velocity Perturbation Score ($\mathcal{VS}_{i}$) for the $i$th object is given by

\[
    d_{i, k} = ||w_{i, k} - w_{ego, k}||^2
\]
\[
    k^{*}_{i} = \argmin \limits_{k \in [0, \cdots, K-1]} d_{i, k} 
\]

\begin{equation}
    \mathcal{VS}_{i} = \begin{cases}
                        -k^{*}_{i} & \text{if $d_{i, k^{*}_{i}} < \tau$} \\
                        -K & \text{otherwise}
    \end{cases}
    \label{eq:VS}
\end{equation}
Here, $w_{ego, k}$ is the $k$th waypoint of the predicted (or perturbed) ego vehicle's trajectory. Similarly, $w_{i,k}$ is the $k$th waypoint of the $i$th non ego vehicle's trajectory. $K$ is the total number of waypoints; $k_i^*$ is the index of the waypoint at which the ego vehicle's trajectory is predicted to be closest to the $i$th vehicle.  If this distance is less than the safety threshold $\tau$, then we assign to the vehicle a score of $-k_i^*$, i.e. a closer index of collision corresponds to a higher score.

\noindent\textbf{Combining the scores:} The final importance score $\mathcal{IS}_{i}$ for a vehicle $i$ is given by the maximum of the two scores: 
\begin{equation}
    \mathcal{IS}_{i} = \max (\mathcal{RS}_{i}, \mathcal{VS}_{i})
    \label{eq:importance score}
\end{equation}
Here, $\mathcal{RS}_{i}$ is the removal score for the $i$th vehicle, and $\mathcal{VS}_{i}$ is the velocity perturbation score for the $i$th vehicle. Both scores are normalized between $0$ and $1$ using their maximum and minimum values across the dataset.

\subsection{Pedestrian Importance Estimation}
\label{subsec:pedestrian_importance}
In the HOIST dataset, the movement patterns of pedestrians are significantly different from vehicles. Vehicle movement is confined to the roads, and they typically adhere to traffic regulations. Pedestrians have more freedom in their movement. They sometimes violate traffic regulations and can even dart out onto the roads. Hence, the mechanisms for estimating importance of vehicles do not necessarily apply to pedestrians. Since pedestrian movement is less constrained, their distance to the ego vehicle is a strong indication for importance. Hence, we ascribe importance to pedestrians based on their L2-distance from the ego vehicle. The importance score ($\mathcal{PS}_{p}$) for a pedestrian $p$ if given by:
\begin{equation}
    \mathcal{PS}_{p} = -||x_{p} - x_{ego}||^{2}
\end{equation}
Here, $x_{p}$ and $x_{ego}$ are the locations of the pedestrian and the ego vehicle, respectively. 

\subsection{Implementation Details}
\label{subsubsec:implementation}
We use LAV~\cite{chen2022learning} to predict the ego vehicle trajectory in our experiments. LAV uses LIDAR point clouds as input for predicting the trajectories of the vehicles. Thus, the counterfactual observations for the Removal Score ($\mathcal{RS}$) are created by removing the points within the 3D bounding box of the concerned objects. We obtain  ground-truth bounding boxes from the simulator, although a 3D bounding box detection module~\cite{vora2020pointpainting} can also be used. 
In addition to the ego vehicle trajectory, LAV can also predict the trajectories of the non-ego vehicles in the scene. However, LAV ignores the objects behind the ego vehicle. We use a constant velocity model for predicting the trajectories of the vehicles ignored by LAV. We average the speed and direction of the vehicles over $5$ previous timesteps. Each trajectory predicted by LAV consists of $K$ waypoints; we set $K = 20$.

\section{Experiments}
\label{sec:experiments}
We evaluate our approach for important object identification against three baselines using the HOIST dataset. Human annotations are used as ground truth. An object is considered important in the ground truth data if it is marked important by at least $\theta_{1}$ annotators. Objects marked important by fewer than $\theta_{2}$ annotators are considered unimportant. Objects marked important by fewer than $\theta_1$ but no less than $\theta_2$ annotators are ignored, i.e., approaches are not penalized for identifying them incorrectly. We do so to make our evaluation robust towards outliers in the dataset. For our experiments, we set $\theta_{1} = 3$ and $\theta_{2} = 2$.

\begin{figure*}
    \centering
    \begin{subfigure}[t]{0.9\linewidth}
        \begin{subfigure}[t]{0.22\linewidth}
            \centering
            \includegraphics[width=\linewidth]{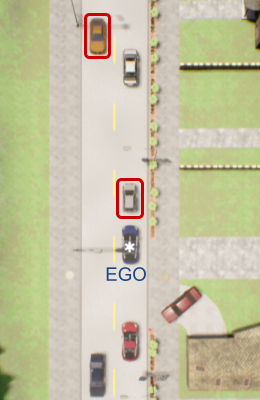}
        \end{subfigure}
        \hfill
        \begin{subfigure}[t]{0.22\linewidth}
            \centering
            \includegraphics[width=\linewidth]{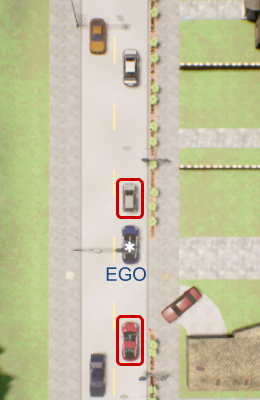}
        \end{subfigure}
        \hfill
        \begin{subfigure}[t]{0.22\linewidth}
            \centering
            \includegraphics[width=\linewidth]{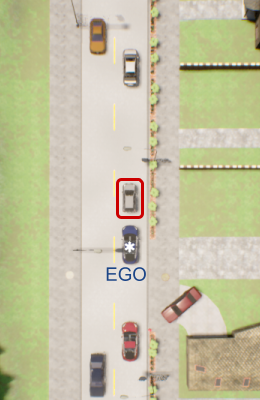}
        \end{subfigure}
        \hfill
        \begin{subfigure}[t]{0.22\linewidth}
            \centering
            \includegraphics[width=\linewidth]{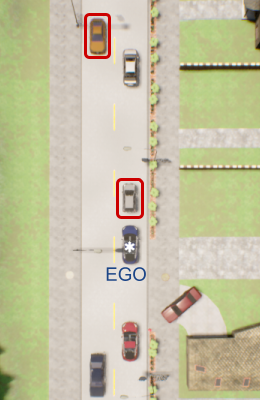}
        \end{subfigure}
    
        \hfill
    
        \begin{subfigure}[t]{0.22\linewidth}
            \centering
            \includegraphics[width=\linewidth]{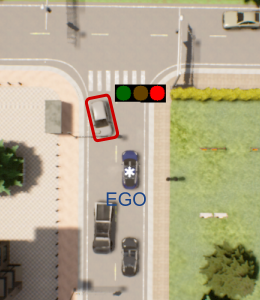}
        \end{subfigure}
        \hfill
        \begin{subfigure}[t]{0.22\linewidth}
            \centering
            \includegraphics[width=\linewidth]{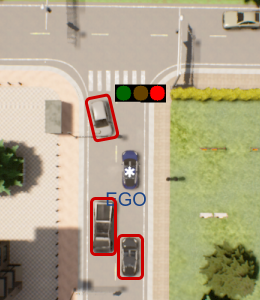}
        \end{subfigure}
        \hfill
        \begin{subfigure}[t]{0.22\linewidth}
            \centering
            \includegraphics[width=\linewidth]{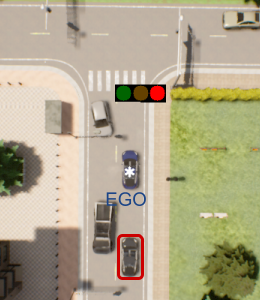}
        \end{subfigure}
        \hfill
        \begin{subfigure}[t]{0.22\linewidth}
            \centering
            \includegraphics[width=\linewidth]{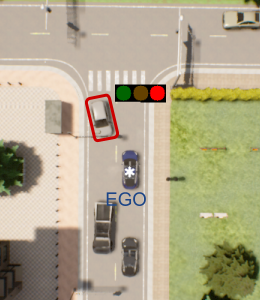}
        \end{subfigure}
    
        \hfill

        \begin{subfigure}[t]{0.22\linewidth}
            \centering
            \includegraphics[width=\linewidth]{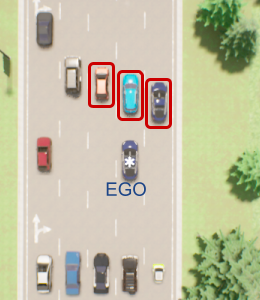}
            \caption{\textbf{Ground-truth (human annotation)}}
        \end{subfigure}
        \hfill
        \begin{subfigure}[t]{0.22\linewidth}
            \centering
            \includegraphics[width=\linewidth]{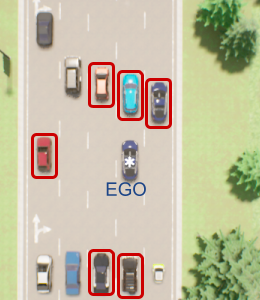}
            \caption{\textbf{Inverse-Distance}}
        \end{subfigure}
        \hfill
        \begin{subfigure}[t]{0.22\linewidth}
            \centering
            \includegraphics[width=\linewidth]{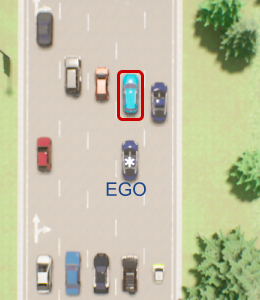}
            \caption{\textbf{PlanT}}
        \end{subfigure}
        \hfill
        \begin{subfigure}[t]{0.22\linewidth}
            \centering
            \includegraphics[width=\linewidth]{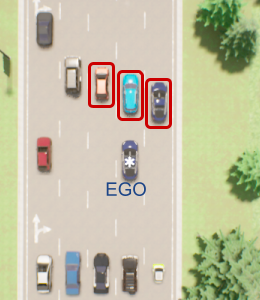}
            \caption{\textbf{Our approach}}
        \end{subfigure}
    \end{subfigure}
  
\setlength{\belowcaptionskip}{-5pt}    
\caption{Qualitative comparison of our approach with the baselines. The objects are marked as  important based on the optimal threshold for the F1 Score for each approach.}
  \label{fig:fig6}
\end{figure*}

\subsection{Comparison with Baselines}
\label{subsec:baselines}

We compare our method with the following baselines for estimating object importance:

\noindent\textbf{Everything Important:} All objects within the scene are considered equally important. We compare our approach to this baseline to highlight the need for the important object identification task.  

\noindent\textbf{Inverse Distance:} Objects are deemed important based on their proximity to the ego vehicle. Specifically, the importance score is the negative L2 distance between the object and the ego vehicle.

\noindent\textbf{PlanT~\cite{Renz2022CORL}:} PlanT is a transformer-based approach for planning in autonomous driving. However, PlanT also outputs relevance scores for vehicles in the form of learned attention weights. We use PlanT's predicted relevance score as the importance score for the objects in our dataset. 

\begin{table}[tb]
    \centering
    
    \begin{tabular}{p{0.35\linewidth}ccc}
        \toprule
        Approach  & AP $\uparrow$ & OT-F1 score $\uparrow$ & OT-Accuracy $\uparrow$ \\
        \midrule
        Everything Important & 0.159  & 0.277 & 90.5 \%  \\ 
        PlanT & 0.572 & 0.594 & 92.7 \% \\
        Inverse Distance & 0.630  & 0.599 & 93.1 \%  \\
        \midrule
        Ours & \textbf{0.710} & \textbf{0.708} & \textbf{94.8} \% \\
        
        \bottomrule
    \end{tabular}
    \setlength{\belowcaptionskip}{-15pt}
    \caption{ Our approach outperforms the baseline approaches with a considerable margin on all metrics.}
    \label{tab:baseline}
    
\end{table}

Table~\ref{tab:baseline} quantitatively compares our approach with the baseline approaches discussed above. We report the Average Precision (AP), the Optimal Threshold F1 Score (OT-F1), and the Optimal threshold Accuracy (OT-Acc) values. The optimal threshold F1 score and accuracy are the highest F1 score and accuracy values across different thresholds. 

The results in Table~\ref{tab:baseline} show that our method outperforms all the baseline approaches across the three metrics. Compared to our approach, the ``Everything Important" approach performs significantly worse ($0.551$ lower AP, $0.431$ lower OT-F1 score, and $4.3$\% worse OT-Accuracy). This shows that attending to every object in the scene is not necessary in order to make safe driving decisions for the ego vehicle.

Our approach also outperforms PlanT on all the metrics ($0.138$ lower AP, $0.114$ lower OT-F1 score, and $2.1$\% worse OT-Accuracy). PlanT considers a very short horizon when reasoning about the state of the ego vehicle and other objects. It only predicts four waypoints into the future for the ego vehicles and reasons about the states of the other objects only for the next timestep. Due to this, PlanT usually assigns a high relevance score to only a few objects nearest to the ego vehicle. Further, it does not consider perturbations in the trajectories of other objects, which can cause it to miss some objects that would be important under perturbations. 

Our approach also outperforms the ``Inverse Distance" baseline ($0.080$ lower AP, $0.109$ lower OT-F1 score, and $1.7\%$ lower OT-Accuracy). This demonstrates that proximity does not always indicate importance. Importance can be better estimated by reasoning about the trajectories of the non-ego vehicles and their influence on the ego vehicle's driving. The qualitative examples in \fig{fig:fig6} shed more light on the reasons behind the superior performance of our approach.

\begin{table}[t]
    \centering
    \begin{tabular}{p{0.3\linewidth}ccc}
        \toprule
        Approach  & AP $\uparrow$ & OT-F1 score $\uparrow$ & OT-Accuracy $\uparrow$ \\
        \midrule
        Object Removal Score & $0.466$ & $0.444$ & $91.7$ \%  \\
        Velocity Perturbations & $0.621$ & $0.670$ & $94.2$ \% \\
        
        \midrule
        Ours & \textbf{0.710} & \textbf{0.708} & \textbf{94.8} \% \\
        \midrule
        Ours w/o Hard Stop & $0.652$  & $0.628$ & $93.9$ \%  \\
        Ours w/o Speed Up & $0.701$ & $0.695$ & $94.5$ \%\\
        Ours w/o Lane Change & $0.565$ & $0.512$ & $93.1$ \%\\
        Ours w/o Ego vehicle Perturbations & $0.635$ & $0.600$ & $93.7$ \%\\
        Ours w/o Index based Weighting & $0.435$ & $0.601$ & $90.5 $ \% \\
        \bottomrule 
    \end{tabular}
    \setlength{\belowcaptionskip}{-20pt}    
    \caption{Ablation experiments demonstrate the impact of the various components and design choices of our method.}
    \label{tab:ablations}
    
\end{table}

\subsection{Ablations}
\label{subsec:ablations}
We perform extensive ablation experiments to highlight the importance of the various components of our approach and justify our design choices. We first ablate over the two counterfactual scenarios, i.e., the removal score ($\mathcal{RS}$) and the velocity perturbation scores ($\mathcal{VS}$).  The results on the top half of Table~\ref{tab:ablations} show that using these scores alone perform worse on all three metrics compared to our method.
This demonstrates the complementary nature of the two scores and the benefit of combining them, as in our method. 
Next, we ablate over the three velocity perturbations. The bottom half of Table~\ref{tab:ablations} shows that removing any of the three velocity perturbations worsens performance. 
Removing the ``lane change" perturbation results in an especially large drop in performance ($0.145$ lower AP). This might be because the ``lane change" perturbation assigns a high importance score to objects moving in lanes adjacent to the ego vehicle. 

Further, we try to understand the benefits of perturbing the ego vehicle trajectory (a defensive driving measure, in case the ego vehicle needs to make an emergency maneuver).
We observe a decrease in performance (drop of $0.075$ in AP) if we do not perform velocity perturbations for the ego vehicle. Finally, we perform an ablation experiment to assess the significance of index based weighting in the velocity perturbation score. We score vehicle importance due to the Velocity Perturbations based on how soon they are predicted to collide with the ego vehicle (Eq.~\ref{eq:VS}). Without index based weighting (treating all vehicles identified as important due to velocity perturbations with equal importance), the performance drops drastically across all metrics (AP drops by $0.357$, OT-F1 score drops by $0.107$, and OT-Accuracy drops by $4.3\%$). Removing index based weighting results in a higher false positive rate as vehicles that influence the ego vehicle's decisions immediately are considered equally important as those which affect the ego vehicle's decisions later.

\subsection{Importance Prediction for Vehicles and Pedestrians}
\label{pedestrian_analysis}
\begin{table}[t]
    \centering
    \begin{tabular}{p{0.35\linewidth}ccc}
        \toprule
        Approach  & AP $\uparrow$ & OT-F1 score $\uparrow$ & OT-Accuracy $\uparrow$ \\(Only Vehicles) &  &  &  \\
        \midrule
        Everything Important & 0.151  & 0.265 & 91.0 \%  \\ 
        PlanT & 0.577 & 0.598 & 93.0 \% \\
        Inverse Distance & 0.614  & 0.593 & 93.2 \%  \\
        \midrule
        $IS$ (Eq~\ref{eq:importance score}, Ours - Vehicles) & \textbf{0.688} & \textbf{0.697} & \textbf{94.8} \% \\
        
        \bottomrule
    \end{tabular}
    \caption{Comparison of our approach with the baseline approaches for \emph{vehicle} importance estimation. Our approach outperforms the baseline approaches across all $3$ metrics}
    \label{tab:baseline_vehicles}
    
\end{table}

In a dynamic traffic scene, vehicles and pedestrians have different movement characteristics, and hence we use different approaches for estimating the importance of vehicles and pedestrians (see Sec~\ref{subsec:pedestrian_importance}). To understand the performance of our approach on vehicles versus pedestrians, we also analyzed each of these categories separately. 

Table~\ref{tab:baseline_vehicles} compares our approach for vehicles only (Sec~\ref{subsec:vehicle_importance}) with the baseline approaches (Sec~\ref{subsec:baselines}). 
We observe similar trends as for the full dataset (vehicles \& pedestrians). Our method outperforms the other baselines on all metrics.

Table~\ref{tab:baseline_pedestrian} compares our approach for pedestrians only (Sec.~\ref{subsec:pedestrian_importance}) against the baseline approaches (Sec.~\ref{subsec:baselines}). As described in Sec.~\ref{subsec:pedestrian_importance}, we use inverse distance for pedestrian importance estimation. 
Using inverse distance outperforms all the other approaches across the three metrics. Its near-perfect AP ($0.996$) shows that pedestrians are deemed important based on their distance from the ego vehicle. PlanT only reasons over a short horizon (Sec~\ref{subsec:baselines}), hence it frequently identifies pedestrians closer to the ego vehicle as important, resulting in a good performance. We also experiment with using our proposed approach for vehicles ($\mathcal{IS}$, Eqn.~\ref{eq:importance score}) to estimate pedestrian importance; this approach also does not perform as well as using inverse distance, and hence we use inverse distance for estimating pedestrian importance in our method. 
While $IS$ is able to identify pedestrians as important when they venture out on the road due to the removal score ($\mathcal{RS}$), it fails to identify the pedestrians on the sidewalks as important.

\begin{table}[t]
    \centering
    \begin{tabular}{p{0.35\linewidth}ccc}
        \toprule
        Approach  & AP $\uparrow$ & OT-F1 score $\uparrow$ & OT-Accuracy $\uparrow$ \\
        (Only Pedestrians) &  &  &  \\
        
        \midrule
        Everything Important & 0.600  & 0.749 & 60.0 \%  \\ 
        PlanT & 0.974 & 0.926 & 90.0 \% \\
        $IS$ (Eq.~\ref{eq:importance score}) & 0.963 & 0.960 & 95.0 \% \\
        \midrule
        Inverse Dist. (Ours - Ped) & \textbf{0.996}  & \textbf{0.978} & \textbf{97.5} \%  \\
        
        \bottomrule
    \end{tabular}
    \caption{Comparison of the baseline approaches for \emph{pedestrian} importance estimation. $\mathcal{IS}$  refers to our approach for \emph{vehicle} importance estimation (Sec.~\ref{subsec:vehicle_importance}, Eqn.~\ref{eq:importance score}). Our approach outperforms all other approaches across all $3$ metrics. }
    \label{tab:baseline_pedestrian}
    
\end{table}

\begin{figure}[t]
  \centering
  \begin{subfigure}[t]{\linewidth}
      \begin{subfigure}[t]{0.47\linewidth}
        \centering
        \includegraphics[width=\linewidth]{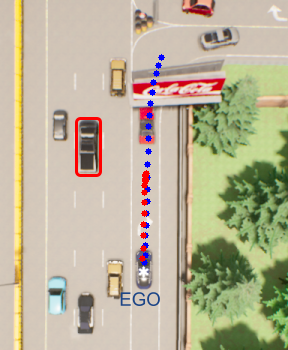}
        \caption{Removing the black truck (red box) incorrectly results in a predicted slow down for the ego vehicle (blue trajectory to red trajectory). }
        \label{fig:fig7a}
      \end{subfigure}
      \hfill
      \begin{subfigure}[t]{0.47\linewidth}
        \centering
         \includegraphics[width=\linewidth]{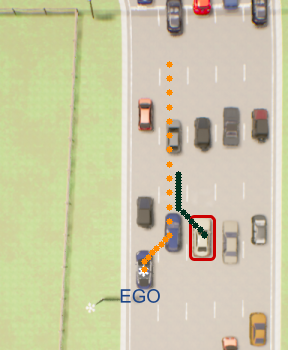}
        \caption{The white car (red box) is incorrectly identified as important. The perturbed trajectories of the ego vehicle and non-ego vehicle are shown in orange and dark green, respectively.}
        \label{fig:fig7b}
      \end{subfigure}
  \end{subfigure}
  \setlength{\belowcaptionskip}{-15pt}    
  \caption{Failure cases of our method.}
  \label{fig:fig7}
\end{figure}

\subsection{Failure Cases}
\label{subsec:failure_cases}

Occasionally, our approach failed to correctly identify important objects. We analyzed the reasons for these failures.

\noindent\textbf{Errors in trajectory prediction:} Our method leverages an autonomous driving agent~\cite{chen2022learning} to predict trajectories of the vehicles in the scene. Hence, our method is prone to errors made by the autonomous driving agent in trajectory prediction. \fig{fig:fig7a} shows a driving scenario under which an incorrect removal score is predicted for vehicles due to an error in the ego vehicle's trajectory prediction. Since the calculation of the Removal Score ($\mathcal{RS}$) involves probing an autonomous driving agent with true and counterfactual driving scenarios, analysing the error cases that arise due to an incorrect removal score can help gain insights about the driving scenarios where autonomous driving agents fail.

\noindent\textbf{Vehicles in far-away lanes:} As a defensive driving measure, we perturb the ego vehicle trajectory along with the other vehicles in the scene. While perturbing the ego vehicle's trajectory overall improves performance (\tbl{tab:ablations}), sometimes this approach marks vehicles with remote chances of collision as important. For example, in \fig{fig:fig7b}, the white vehicle is identified as important, even though the chances of it colliding with the ego vehicle are slim.


\section{Conclusion}
\label{sec:conclusion}
We focus on the task of important object identification in the context of driving. We define important objects as the ones a driver would watch out for in order to drive the ego vehicle safely. This definition of importance is better suited for the development of situational awareness support systems than prior works which only consider vehicles currently influencing the ego-vehicles trajectory. We put forth a new public dataset named HOIST, which contains human-annotated importance labels for vehicles and pedestrians in simulated driving scenarios. Unlike prior works, HOIST consists of multi-modal sensor data used by state-of-the-art autonomous driving agents. We further contribute a novel approach that relies on counterfactual reasoning to identify object importance within dynamic traffic scenes. Our approach outperforms various baselines for the task of important object identification on HOIST. We hope that our work leads the way toward efficient, robust, and explainable algorithms for the task of object importance estimation for driving scenarios.

\bibliographystyle{ieeetr}
\bibliography{root}

\begin{thebibliography}{10}

\bibitem{wu2022toward}
T.~Wu, E.~Sachdeva, K.~Akash, X.~Wu, T.~Misu, and J.~Ortiz, ``Toward an adaptive situational awareness support system for urban driving,'' in {\em 2022 IEEE Intelligent Vehicles Symposium (IV)}, pp.~1073--1080, IEEE, 2022.

\bibitem{biswas2022mitigating}
A.~Biswas, B.~A. Pardhi, C.~Chuck, J.~Holtz, S.~Niekum, H.~Admoni, and A.~Allievi, ``Mitigating causal confusion in driving agents via gaze supervision,'' in {\em Aligning Robot Representations with Humans workshop @ Conference on Robot Learning}, 2022.

\bibitem{palazzi2018predicting}
A.~Palazzi, D.~Abati, F.~Solera, R.~Cucchiara, {\em et~al.}, ``Predicting the driver's focus of attention: the dr (eye) ve project,'' {\em IEEE transactions on pattern analysis and machine intelligence}, vol.~41, no.~7, pp.~1720--1733, 2018.

\bibitem{xia2019predicting}
Y.~Xia, D.~Zhang, J.~Kim, K.~Nakayama, K.~Zipser, and D.~Whitney, ``Predicting driver attention in critical situations,'' in {\em Computer Vision--ACCV 2018: 14th Asian Conference on Computer Vision, Perth, Australia, December 2--6, 2018, Revised Selected Papers, Part V 14}, pp.~658--674, Springer, 2019.

\bibitem{baee2021medirl}
S.~Baee, E.~Pakdamanian, I.~Kim, L.~Feng, V.~Ordonez, and L.~Barnes, ``Medirl: Predicting the visual attention of drivers via maximum entropy deep inverse reinforcement learning,'' in {\em Proceedings of the IEEE/CVF international conference on computer vision}, pp.~13178--13188, 2021.

\bibitem{pal2020looking}
A.~Pal, S.~Mondal, and H.~I. Christensen, ``" looking at the right stuff"-guided semantic-gaze for autonomous driving,'' in {\em Proceedings of the IEEE/CVF Conference on Computer Vision and Pattern Recognition}, pp.~11883--11892, 2020.

\bibitem{fang2019dada}
J.~Fang, D.~Yan, J.~Qiao, J.~Xue, H.~Wang, and S.~Li, ``Dada-2000: Can driving accident be predicted by driver attentionƒ analyzed by a benchmark,'' in {\em 2019 IEEE Intelligent Transportation Systems Conference (ITSC)}, pp.~4303--4309, IEEE, 2019.

\bibitem{kotseruba2021behavioral}
I.~Kotseruba and J.~K. Tsotsos, ``Behavioral research and practical models of drivers' attention,'' {\em arXiv preprint arXiv:2104.05677}, 2021.

\bibitem{white2010blind}
C.~B. White and J.~K. Caird, ``The blind date: The effects of change blindness, passenger conversation and gender on looked-but-failed-to-see (lbfts) errors,'' {\em Accident Analysis \& Prevention}, vol.~42, no.~6, pp.~1822--1830, 2010.

\bibitem{crundall2012some}
D.~Crundall, P.~Chapman, S.~Trawley, L.~Collins, E.~Van~Loon, B.~Andrews, and G.~Underwood, ``Some hazards are more attractive than others: Drivers of varying experience respond differently to different types of hazard,'' {\em Accident Analysis \& Prevention}, vol.~45, pp.~600--609, 2012.

\bibitem{biswas2023characterizing}
A.~Biswas and H.~Admoni, ``Characterizing drivers’ peripheral vision via the functional field of view for intelligent driving assistance,'' in {\em 2023 IEEE Intelligent Vehicles Symposium (IV)}, pp.~1--8, IEEE, 2023.

\bibitem{gao2019goal}
M.~Gao, A.~Tawari, and S.~Martin, ``Goal-oriented object importance estimation in on-road driving videos,'' in {\em 2019 International Conference on Robotics and Automation (ICRA)}, pp.~5509--5515, IEEE, 2019.

\bibitem{zhang2020interaction}
Z.~Zhang, A.~Tawari, S.~Martin, and D.~Crandall, ``Interaction graphs for object importance estimation in on-road driving videos,'' in {\em 2020 IEEE International Conference on Robotics and Automation (ICRA)}, pp.~8920--8927, IEEE, 2020.

\bibitem{li2022important}
J.~Li, H.~Gang, H.~Ma, M.~Tomizuka, and C.~Choi, ``Important object identification with semi-supervised learning for autonomous driving,'' in {\em 2022 International Conference on Robotics and Automation (ICRA)}, pp.~2913--2919, IEEE, 2022.

\bibitem{endsley1995toward}
M.~R. Endsley, ``Toward a theory of situation awareness in dynamic systems,'' {\em Human factors}, vol.~37, no.~1, pp.~32--64, 1995.

\bibitem{chitta2022transfuser}
K.~Chitta, A.~Prakash, B.~Jaeger, Z.~Yu, K.~Renz, and A.~Geiger, ``Transfuser: Imitation with transformer-based sensor fusion for autonomous driving,'' {\em IEEE Transactions on Pattern Analysis and Machine Intelligence}, 2022.

\bibitem{dosovitskiy2017carla}
A.~Dosovitskiy, G.~Ros, F.~Codevilla, A.~Lopez, and V.~Koltun, ``Carla: An open urban driving simulator,'' in {\em Conference on robot learning}, pp.~1--16, PMLR, 2017.

\bibitem{chen2022learning}
D.~Chen and P.~Kr{\"a}henb{\"u}hl, ``Learning from all vehicles,'' in {\em Proceedings of the IEEE/CVF Conference on Computer Vision and Pattern Recognition}, pp.~17222--17231, 2022.

\bibitem{li2020make}
C.~Li, S.~H. Chan, and Y.-T. Chen, ``Who make drivers stop? towards driver-centric risk assessment: Risk object identification via causal inference,'' in {\em 2020 IEEE/RSJ International Conference on Intelligent Robots and Systems (IROS)}, pp.~10711--10718, IEEE, 2020.

\bibitem{li2023droid}
C.~Li, S.~H. Chan, and Y.-T. Chen, ``Droid: Driver-centric risk object identification,'' {\em IEEE transactions on pattern analysis and machine intelligence}, 2023.

\bibitem{vemula2018social}
A.~Vemula, K.~Muelling, and J.~Oh, ``Social attention: Modeling attention in human crowds,'' in {\em 2018 IEEE international Conference on Robotics and Automation (ICRA)}, pp.~4601--4607, IEEE, 2018.

\bibitem{li2021rain}
J.~Li, F.~Yang, H.~Ma, S.~Malla, M.~Tomizuka, and C.~Choi, ``Rain: Reinforced hybrid attention inference network for motion forecasting,'' in {\em Proceedings of the IEEE/CVF International Conference on Computer Vision}, pp.~16096--16106, 2021.

\bibitem{li2021spatio}
J.~Li, H.~Ma, Z.~Zhang, J.~Li, and M.~Tomizuka, ``Spatio-temporal graph dual-attention network for multi-agent prediction and tracking,'' {\em IEEE Transactions on Intelligent Transportation Systems}, vol.~23, no.~8, pp.~10556--10569, 2021.

\bibitem{Renz2022CORL}
K.~Renz, K.~Chitta, O.-B. Mercea, A.~S. Koepke, Z.~Akata, and A.~Geiger, ``Plant: Explainable planning transformers via object-level representations,'' in {\em Conference on Robotic Learning (CoRL)}, 2022.

\bibitem{aksoy2020see}
E.~Aksoy, A.~Yaz{\i}c{\i}, and M.~Kasap, ``See, attend and brake: An attention-based saliency map prediction model for end-to-end driving,'' {\em arXiv preprint arXiv:2002.11020}, 2020.

\bibitem{xia2019driver}
Y.~Xia, {\em Driver eye movements and the application in autonomous driving}.
\newblock University of California, Berkeley, 2019.

\bibitem{xia2020periphery}
Y.~Xia, J.~Kim, J.~Canny, K.~Zipser, T.~Canas-Bajo, and D.~Whitney, ``Periphery-fovea multi-resolution driving model guided by human attention,'' in {\em Proceedings of the IEEE/CVF Winter Conference on Applications of Computer Vision}, pp.~1767--1775, 2020.

\bibitem{ziebart2008maximum}
B.~D. Ziebart, A.~L. Maas, J.~A. Bagnell, A.~K. Dey, {\em et~al.}, ``Maximum entropy inverse reinforcement learning.,'' in {\em Aaai}, vol.~8, pp.~1433--1438, Chicago, IL, USA, 2008.

\bibitem{shao2023reasonnet}
H.~Shao, L.~Wang, R.~Chen, S.~L. Waslander, H.~Li, and Y.~Liu, ``Reasonnet: End-to-end driving with temporal and global reasoning,'' in {\em Proceedings of the IEEE/CVF Conference on Computer Vision and Pattern Recognition}, pp.~13723--13733, 2023.

\bibitem{shao2022interfuser}
H.~Shao, L.~Wang, R.~Chen, H.~Li, and Y.~Liu, ``Safety-enhanced autonomous driving using interpretable sensor fusion transformer,'' {\em arXiv preprint arXiv:2207.14024}, 2022.

\bibitem{wu2022trajectory}
P.~Wu, X.~Jia, L.~Chen, J.~Yan, H.~Li, and Y.~Qiao, ``Trajectory-guided control prediction for end-to-end autonomous driving: A simple yet strong baseline,'' {\em Advances in Neural Information Processing Systems}, vol.~35, pp.~6119--6132, 2022.

\bibitem{vora2020pointpainting}
S.~Vora, A.~H. Lang, B.~Helou, and O.~Beijbom, ``Pointpainting: Sequential fusion for 3d object detection,'' in {\em Proceedings of the IEEE/CVF conference on computer vision and pattern recognition}, pp.~4604--4612, 2020.

\end{thebibliography}
\end{document}